\documentclass[10pt,twocolumn,letterpaper]{article}

\usepackage{iccv}
\usepackage{times}
\usepackage{epsfig}
\usepackage{graphicx}
\usepackage{amsmath}
\usepackage{amssymb}
\usepackage{comment}
\usepackage{float}
\usepackage{verbatim}
\usepackage{textcomp}
\usepackage{booktabs}
\usepackage[utf8]{inputenc}
\usepackage[english]{babel}

\usepackage[dvipsnames]{xcolor}

\definecolor{pink}{cmyk}{0, 0.7808, 0.4429, 0.1412}
\usepackage[pagebackref=true,breaklinks=true,letterpaper=true,colorlinks,bookmarks=false]{hyperref}

\iccvfinalcopy % *** Uncomment this line for the final submission

\def\iccvPaperID{3786} % *** Enter the ICCV Paper ID here

\ificcvfinal\fi

\begin{document}
	
	%%%%%%%%% TITLE
	\title{TRB: A Novel Triplet Representation for Understanding 2D Human Body}
	\vspace{-5mm}
	\author{Haodong Duan$^{1}$, KwanYee Lin$^{2}$, Sheng Jin$^{2}$, Wentao Liu$^{2}$, Chen Qian$^{2}$, Wanli Ouyang$^{3}$\\
		$^{1}$CUHK-Sensetime Joint Lab, $^{2}$SenseTime Group Limited\\ $^{3}$The University of Sydney, SenseTime Computer Vision Research Group, Australia
	}
	
	\maketitle
	%\thispagestyle{empty}
	
	%%%%%%%%% ABSTRACT
	
	%%%%%%%%% ABSTRACT
	\begin{abstract}
		Human pose and shape are two important components of 2D human body. However, how to efficiently represent both of them in images is still an open question. In this paper, we propose the \textbf{Triplet Representation for Body} (TRB) --- a compact 2D human body representation, with skeleton keypoints capturing human pose information and contour keypoints containing human shape information. TRB not only preserves the flexibility of skeleton keypoint representation, but also contains rich pose and human shape information. Therefore, it promises broader application areas, such as human shape editing and conditional image generation. 
		We further introduce the challenging problem of TRB estimation, where joint learning of human pose and shape is required. We construct several large-scale TRB estimation datasets, based on popular 2D pose datasets: LSP, MPII, COCO. To effectively solve TRB estimation, we propose a two-branch network (TRB-net) with three novel techniques, namely X-structure (Xs), Directional Convolution (DC) and Pairwise Mapping (PM), to enforce multi-level message passing for joint feature learning. We evaluate our proposed TRB-net and several leading approaches on our proposed TRB datasets, and demonstrate the superiority of our method through extensive evaluations.
		
	\end{abstract}
	
	\vspace{-3mm}
	%%%%%%%%% BODY TEXT
	\section{Introduction}
	
	A comprehensive 2D human body representation should capture both human pose and shape information. Such representation is promising for applications beyond plain keypoint localization, such as graphics and human-computer interaction. However, how to establish such 2D body representation is still an open problem. Current mainstream 2D human body representations are not able to simultaneously capture both information. Skeleton keypoint based representation~\cite{andriluka20142d,johnson2010clustered,lin2014microsoft} well captures human poses. However, such representation loses the 2D human shape information which is essential for human body understanding. Pixel-wise human parsing representations~\cite{liang2015human,chen2014detect,gong2017look} contain 2D human shape cues. However, such kinds of representations lack accurate keypoint localization information, since all pixels in one part share the same semantic label. Meanwhile, they are inflexible to manipulate and costly to label. This paper aims at discovering a new representation for more comprehensive understanding of the human body. To this end, a novel Triplet Representation for Body (TRB) is introduced. It consists of skeleton and contour keypoint representations, capturing both accurate pose localization and rich semantic human shape information simultaneously, while preserving its flexibility and simplicity.  
	
	Since there exists no dataset to quantitatively evaluate TRB estimation, we propose several challenging TRB datasets based on three pose estimation datasets (LSP~\cite{johnson2010clustered}, MPII~\cite{andriluka20142d} and COCO~\cite{lin2014microsoft}). We quantitatively evaluate the performance of several state-of-the-art 2D skeleton-based keypoint detectors on the proposed TRB datasets. Our experiments indicate that they are not able to effectively solve the more challenging TRB estimation tasks, which require the approaches to not only understand the concept of  human pose and human shape simultaneously, but also exploit the underlying relationship between them.
	
	For effective representation learning, we design a two-branch multi-task framework called TRB-Net, which jointly solves skeleton keypoint estimation and contour keypoint estimation. These two tasks are closely related and will promote each other. Therefore, we design a message passing block to enable information exchange. The message received from the other branch will act as guidance for the current branch to produce finer estimation results. Since feature maps from the two branches have different patterns, spatial feature transformation is necessary for feature alignment and more effective message passing scheme. Therefore, we propose a task-specific directional convolution operator to exploit the inside-out and outside-in spatial relationship between skeleton and contour feature maps. To prevent inconsistent predictions for skeleton and contour branch, we explicitly enforce pairwise mapping constraints. With these techniques, we boost the TRB estimation performance beyond state-of-the-arts. Error is reduced by 13.3\% and 15.1\% for skeleton and contour keypoint estimation respectively (Sec.~\ref{ablation_part}).
	
	Our contributions are three-fold: $(1)$ We propose the novel Triplet Representation for Body (TRB), which embodies both human pose and shape information. We apply TRB to the conditional image generation task, and show its effectiveness in handling pose/shape guided image generation and human shape editing. $(2)$ We introduce a challenging TRB estimation task, establish a benchmark and evaluate various mainstream pose estimation approaches in the context of TRB estimation. $(3)$ We design the TRB-net which jointly learns the skeleton and contour keypoint representation. Three techniques are proposed for effective message passing and feature learning. Extensive experiments show the effectiveness of our proposed methods.
	
	%-------------------------------------------------------------------------
	
	\section{Related Work}
	\begin{comment}
	{\color{red}
	\textbf{2D human body representation} 
	Earlier works on 2D human body representation mainly rely on pictorial structure model~\cite{felzenszwalb2005pictorial,andriluka2009pictorial}, which uses a set of rectangles with center coordinates, scale and orientation to represent the human body. Such kinds of representation is limited by the shape of rectangular, leading to low-quality localization, and high ambiguity on the boundary. Recently, skeleton keypoints~\cite{andriluka20142d,johnson2010clustered,lin2014microsoft} become the most popular human pose representation for its simplicity, feasibility and effectiveness. The articulated keypoint representation has rich semantic meaning but lacks human shape or contour information. Pixel-wise labeling may include semantic human parsing representation~\cite{liang2015human,chen2014detect,gong2017look} and DensePose~\cite{alp2018densepose} 3D surface-based representation. Such kinds of representations better preserve the human shape, while lacking accurate keypoint localization information and have high labeling cost. In this work, we extend the widely used skeleton keypoint representation and propose a novel triplet representation for 2D human body understanding. It not only captures accurate articulated localization information, but also contains rich semantic human shape information.  }
	\end{comment}

	\textbf{Human body representation. } 
	On 2D human body representation, Pictorial structures (PS)~\cite{felzenszwalb2005pictorial,andriluka2009pictorial} was most popular in the early stage, which uses a set of rectangles to represent articulated human limbs. Deformable structure~\cite{zuffi2012pictorial} and contour people~\cite{freifeld2010contour} further improved PS by using dense contour points instead of rigid rectangles, leading to better shape fitting. However, these representations are too complicated to annotate and can only be optimized by energy minimization. Recently, researchers used skeleton keypoints~\cite{andriluka20142d,johnson2010clustered,lin2014microsoft} as 2D human pose representation for its simplicity, feasibility and effectiveness. The articulated keypoint representation has rich semantic meaning but lacks human shape or contour information. Pixel-wise labeling may include semantic human parsing representation~\cite{liang2015human,chen2014detect,gong2017look} and DensePose~\cite{alp2018densepose} 3D surface-based representation which preserve the human shape. However, those representations lack accurate keypoint localization information and have high labeling cost. There are other 3D body models, including SMPL~\cite{loper2015smpl} and SCAPE~\cite{anguelov2005scape}, which can represent both pose and shape. However, 3D human annotation in the wild is hard to obtain, which makes it difficult to exploit these models for 2D human understanding and 2D image editing. In this work, we extend the widely used skeleton keypoint representation and propose a novel triplet representation for 2D human body understanding, which not only captures accurate articulated localization information, but also contains rich semantic human shape information.  
	
	\textbf{Human pose estimation} With the flourishing of deep learning, CNN-based models have dominated both single-person and multi-person pose estimation problems~\cite{tompson2015efficient, tompson2014joint, chu2016crf, cao2016realtime, yang2016end, papandreou2017towards, pishchulin2016deepcut, newell2017associative, jin2019multi}. DeepPose~\cite{toshev2014deeppose} first proposes to leverage CNN features to directly regress human keypoint coordinates. However, such keypoint coordinate representation is highly abstract and is ineffective to exploit visual cues, which encourages researchers to explore better representation~\cite{newell2016stacked,wei2016convolutional,chu2017multi,yang2017learning}. CPM~\cite{wei2016convolutional} explores a sequential composition of convolutional architectures which directly operate on belief maps from previous stages to implicitly learn spatial relationship for human pose estimation. Stacked hourglass~\cite{newell2016stacked} uses repeated bottom-up, top-down processing in conjunction with intermediate supervision on heatmaps to improve the performance. Chu \etal propose a multi-context attention~\cite{chu2017multi} scheme to utilize multi-level structural information and achieve more robust and accurate prediction. Yang \etal propose to construct Pyramid Residual Module at the bottom of the stacked hourglass network, to enhance the scale in-variance of deep models by learning feature pyramids~\cite{yang2017learning}. 
	These methods only focus on 2D skeleton keypoint localization, while we extend it to both skeleton and contour keypoint estimation for better understanding of 2D human body. 
	
	\textbf{Multi-task learning in human analysis}
	Multi-task learning~\cite{misra2016cross, zhang2014facial} is widely used in human analysis, knowledge transferring between different tasks can benefit both. In~\cite{gkioxari2018detecting}, the action detector, object detector and HOI classifier are jointly trained to predict human object relationship accurately.  In~\cite{Nie_2018_CVPR, nie2018mutual}, dynamic convolution was used for message passing between two tasks. In dynamic convolution, the dynamic convolution parameters was learned from one task, while the convolution was performed on the other task. In~\cite{luvizon20182d}, pose estimation was jointly trained with downstream application action recognition.  Considering natural spatial relationship between human skeleton estimation and contour estimation, we further proposed three domain knowledge transferring modules beyond plain multi-task training. The proposed modules fit the nature of the tasks well and make message passing between 2 tasks more efficient, as demonstrated in our experiments.
	
	\section{Triplet Representation for Body (TRB)}
	\label{defn}
	
	\begin{figure}
		\centering
		\includegraphics[page=1,width=0.42\textwidth]{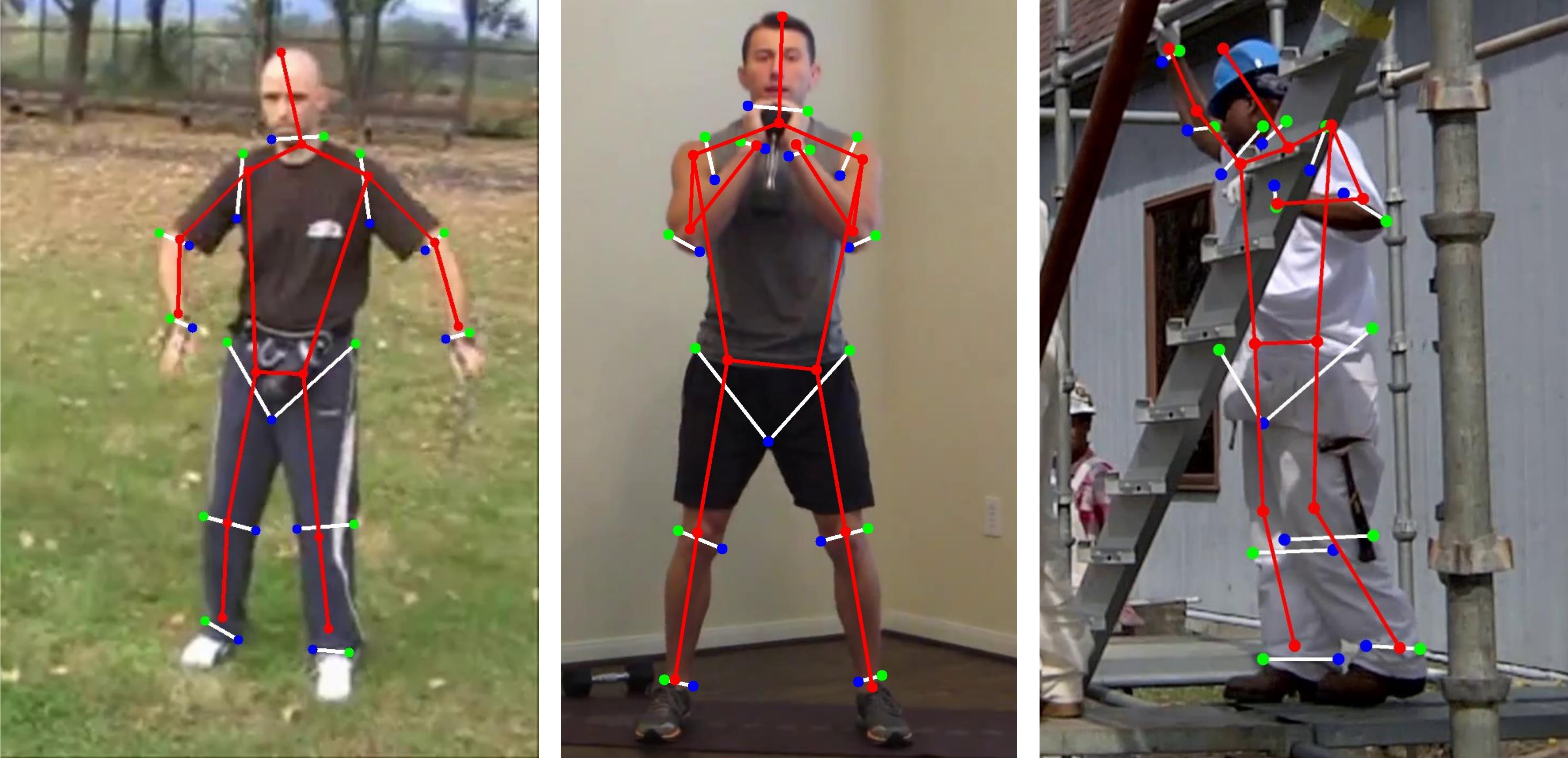}
		\caption{\label{sup_label}\small{\textbf{Example Annotations.} We visualize some samples of our labeled images. Red dots denote the \emph{skeleton} keypoints. Green dots denote lateral \emph{contour} keypoints. Blue dots denote medial \emph{contour} keypoints. Human skeleton is demonstrated using red lines while contour keypoints belong to the same triplet are connected using white lines.}}
	\end{figure}
	
	\subsection{Concept of TRB}
	
	Recently, skeleton keypoint representation has become the most popular human body representation, because of its simplicity, feasibility and  effectiveness. However, such kind of representation fails to capture the human shape information. Due to lack of shape information, the potential of 2D human body representation in many real-world applications is not fully explored. Some suggest using pixel-level human parsing annotation to preserve human shape. However, the accurate localization information and keypoint semantic information is missing. 
	
	In this work, we propose a novel representation for both 2D human pose and shape, called Triplet Representation for Body (\textbf{TRB}). We design a compact representation, where contour keypoints located on the human boundary represent the human shape, skeleton keypoints preserve the human articulated structure. Our proposed representation is more feasible and easier to label, while preserving both the rich human boundary information and accurate keypoint localization information. Specifically, a human body is represented by a set of triplet descriptors. Each keypoint triplet consists of a skeleton keypoint and two nearby contour keypoints located on the human boundary. We classify the contour keypoints into medial contour verses lateral contour keypoints to avoid semantic ambiguity. As shown in Fig.~\ref{sup_label}, in each triplet, one of the two contour keypoints is located on the medial side (Blue), while the other is located on the lateral side (Green). The two contour keypoints are pre-defined with clear and explicit semantic meaning, while preserving strong visual evidence. 
	
	By introducing TRB, we unify the representation of 2D pose and shape in an efficient way, which benefits the downstream applications such as human parsing, human shape editing, etc.. Moreover, as a side-effect, the extra contour keypoints also provide boundary cues for skeleton keypoint localization, and vice versa.

	\subsection{TRB Estimation Task, Dataset and Evaluation}
	In this section, we introduce the TRB Estimation task. TRB estimation task is to estimate the whole set of TRB keypoint triplets (including both skeleton and contour keypoints) for each person from a single RGB image. It is more challenging than previous 2D skeleton keypoint localization tasks, since it requires a more comprehensive understanding of the human body, including pose, shape and their relationship.
	
\begin{figure}
		\centering
		\includegraphics[page=1,width=0.42\textwidth]{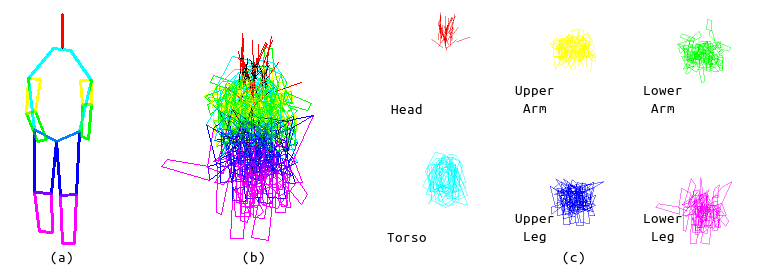}
		\caption{\label{lbvis}\small{\textbf{Shape visualization.} (a) Color coding of body parts. (b)Human contour variability in MPII (Random 40 people aligned with the same center).
(c)Human contour variability for each part.
		}}
	\end{figure}

	We build three TRB datasets based on MPII~\cite{andriluka20142d}, LSP~\cite{johnson2010clustered} and COCO~\cite{lin2014microsoft}, denoted as MPII\_trb, LSP\_trb and COCO\_trb respectively. MPII and LSP are popular single person pose estimation dataset, which contain around 40K and 12K person annotated poses respectively. We annotate the contour keypoints on all the train-val data of MPII and LSP, and build MPII\_trb and LSP\_trb which contain a whole set of skeleton and contour triplets. COCO is a much larger dataset with around 150K annotated people. For COCO, we randomly annotated half of its train-val data to form COCO\_trb dataset. Fig.\ref{lbvis} displays some TRB annotations on MPII dataset, the highly variable human shape emphasized the importance of capturing human shape in 2D human representation.

	The proposed TRB datasets are compatible with their corresponding 2D pose datasets. For example, occlusion cases were dealt in accordance with the labeling protocol of the 2D pose dataset. In specific, for MPII and LSP datasets, all occluded contour keypoints are labeled with estimated positions. For COCO, the occluded ones are annotated only if the corresponding skeleton keypoints are annotated. TRB estimation task employs the same evaluation metrics as the common 2D skeleton keypoint estimation task.
	
	\begin{figure*}
		\centering
		\includegraphics[page=1,width=0.82\textwidth]{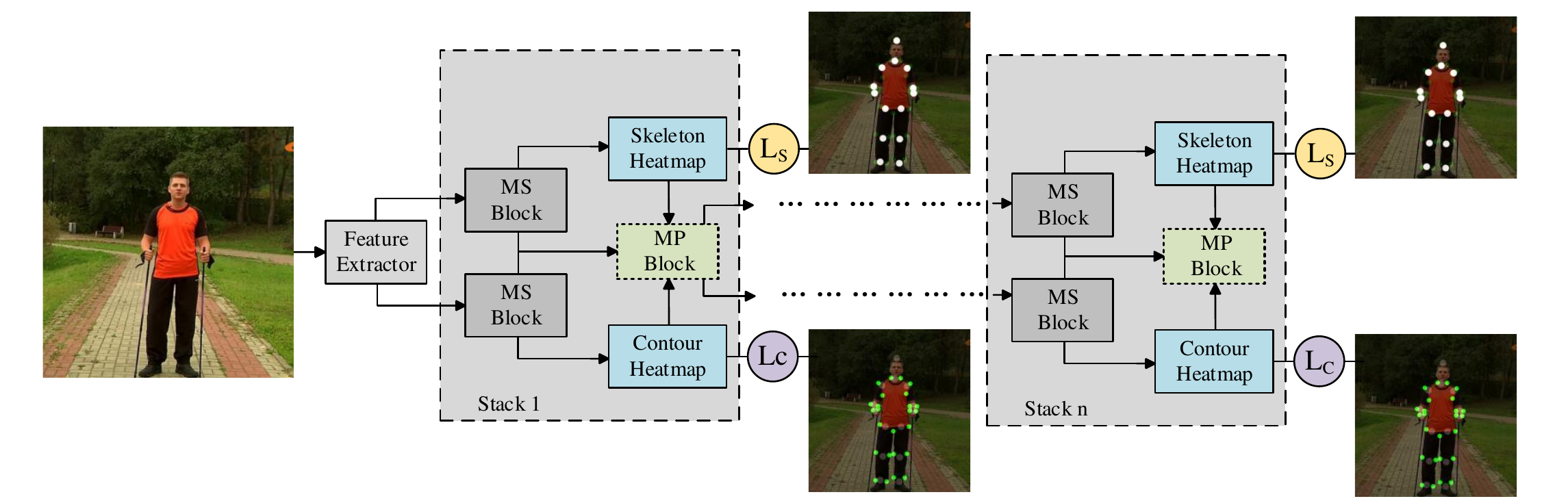}
		\caption{\label{framework}\small{\textbf{Framework.} The framework of our proposed TRB-net for joint contour and skeleton keypoint estimation. The message passing (\textbf{MP}) blocks represent plug-in modules to enhance message passing between branches, including X-structure (\textbf{Xs}), Directed Convolution (\textbf{DC}) and Pairwise Mapping (\textbf{PM}) module. $L_S$, $L_C$ represent skeleton and contour loss respectively, which are used as intermediate supervision.
		}}
	\end{figure*}
	
	\section{TRB-Net}
	
\begin{figure*}
		\centering
		\includegraphics[page=1,width=0.82\textwidth]{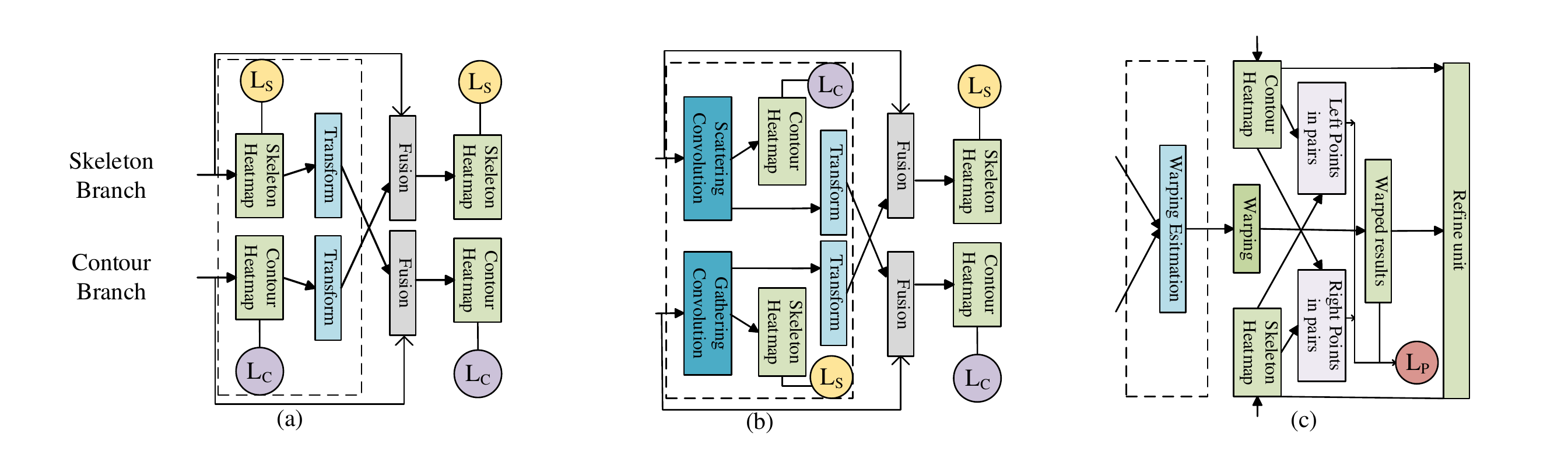}
		\caption{\label{new_modules}\small{\textbf{Message Passing modules.} Three plug-in message passing modules are illustrated. (a) denotes the X-structured (\textbf{Xs}) message passing block, in which heatmaps produced by one branch are passed to the other branch for information exchange. (b) represents Directed Convolution unit (\textbf{DC}), in which Scattering and Gathering convolutions are utilized for efficient message passing. (c) denotes the Pairwise Mapping unit (\textbf{PM}), where pairwise constraints are introduced to improve the consistency of the predictions and Refine unit is employed to obtain finer results. The dashed box denotes the part to fit in MP Block in Fig.~\ref{framework}. }}
	\end{figure*}

	We propose the TRB-Net to jointly solve the problem of skeleton and contour keypoint estimation. Our overall framework is illustrated in Fig.~\ref{framework}. TRB-Net follows the widely used multi-stack framework to produce coarse to fine prediction. In each stack, the model consists of a skeleton branch for the skeleton keypoint prediction and a contour branch for contour landmark prediction. In both branches, multi-scale feature extraction block (MS Block) is utilized for effective feature learning. We also propose the X-structured Message Passing Block (MP Block) to enforce the mutual interaction between these two branches (see Sec~\ref{Xstr}, Fig~\ref{new_modules}(a)). And for effective spatial transformation, we further design a novel convolution operation, namely Directional Convolution (DC), which encourages message passing on specific direction (see Sec~\ref{dirconv}, Fig~\ref{new_modules}(b)). Finally, we propose the Pairwise Mapping (PM) module to enforce the consistency of the skeleton and contour predictions (Sec~\ref{sec_pairwise_mapping}, Fig~\ref{new_modules}(c)). 
	
	We add several intermediate supervision ($L_S$, $L_C$ and $L_P$) to train the model. $L_S$, $L_C$, $L_P$ represent skeleton loss, contour loss and pairwise loss respectively. $L_S$ and $L_C$ measure the $L2$ distances between the predicted and ground truth heatmaps. The pairwise loss measures the inconsistency of pairwise mapping (Sec~\ref{sec_pairwise_mapping}).
	
	\subsection{X-Structured Message Passing Block} 
	\label{Xstr}
	
	As stated above, our model consists of a skeleton branch and a contour branch. Considering the strong spatial relationship between skeleton and contour keypoints, we design a X-Structured message passing (MP) block to explicitly strengthen the information exchange (see Fig.\ref{new_modules}(a)). As shown in Fig.\ref{framework}, the X-structured module enables message passing at different feature learning stages, to obtain better feature representation for both tasks. By introducing the X-structured MP block, the skeleton branch is able to get guidance from the contour branch for more accurate localization, and vice versa. Since keypoint heatmaps contain clear semantic meanings and rich spatial information, it can be used as a strong prior for keypoint localization. Therefore, we choose transformed heatmaps as the messages to be transferred between branches. Take the skeleton branch for example, the coarse skeleton heatmaps are first mapped to the space of the contour with a Transform module. Then the transformed heatmaps are sent to the contour branch as messages. Finally, coarse contour heatmaps and the received messages are adaptively fused to produce finer contour heatmap predictions. In our implementation, the Transform module performs feature mapping with 1x1 convolution, and the Fusion module concatenates two source heatmaps and fuse them with 1x1 convolution.
	
	\begin{figure*}
		\centering
		\includegraphics[page=1,width=0.82\textwidth]{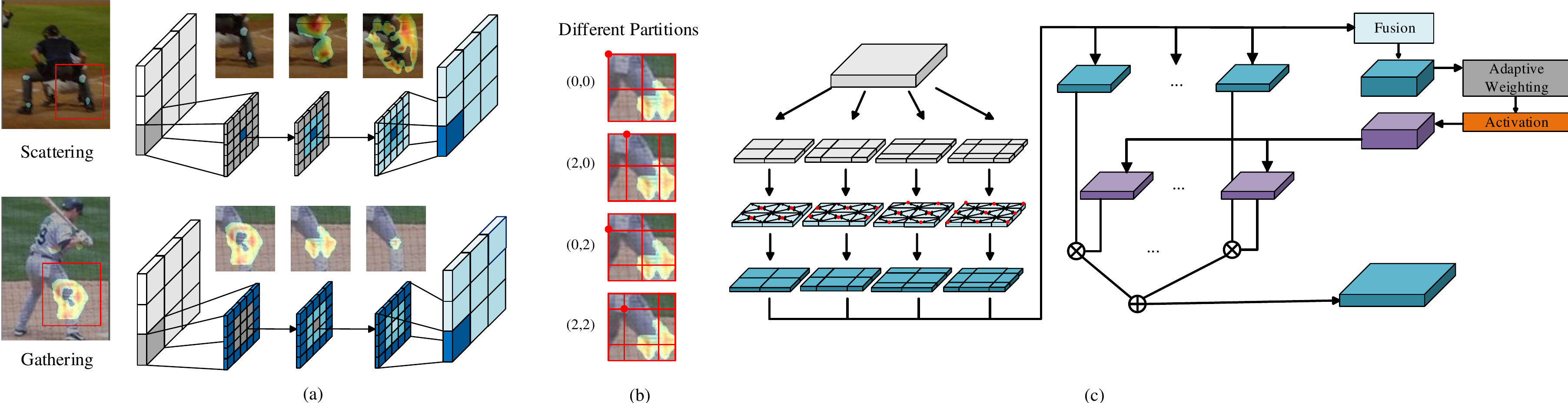}  
		\caption{\label{block_conv}\small{\textbf{Directional Convolution}. (a) Directional convolution on a 5x5 block. Updating is conducted in inside-out or outside-in order for Scattering and Gathering convolutions. Related feature maps are visualized, where the already updated areas are colored in blue. (b) Four different partitions (red lines) for a 8x8 feature map with different grid partition point. Red dots denote grid partition points, whose relative positions are written on the left. (c) Directional convolution is conducted in parallel on different partitions of feature maps. The relative center of Scattering and Gathering convolutions are marked by red dots. The results are fused using the adaptive self-attention scheme. 
		}}
	\end{figure*}
	
	\subsection{Directional Convolution}
	\label{dirconv}
	
	In the previous section, we use a simple Transform module to map the skeleton heatmaps to the contour space. However, activation in skeleton branch is often concentrates on skeleton, while activation in contour branch often distributes around skeleton (see Fig.~\ref{block_conv}). A task-specific local spatial transformation is needed to better align the activation maps before message passing. To this end, we design a novel iterative convolution operator, called \emph{directional convolution}. It enables oriented spatial transformation at the feature level explicitly and efficiently. 
	
	We first consider an oriented feature-fusion problem defined on a small sub-region and attempt to address this sub-problem. A directional convolution operator $T$ consists of $K$ iterative convolution steps. A valid directional convolution should meet the following requirements. $(1)$ in each step, only a set of pixels are updated. $(2)$ After the last iteration, all pixels should be updated once and only once. A sequence of characteristic functions $F = \{F_{k}\}_{k=1}^{K}$ is defined to control the updating order of pixels. The input of function $F_k$ is the position of a pixel on heatmap while the output is 1 or 0. $F_k$ denotes whether to update a pixel in the $k_{th}$ iteration. In specific, we only update the area where $F_k=1$ and keep other areas fixed. The updating of the $i_{th}$ iteration can be formulated as: 
	
	\vspace{-3mm}
	\begin{equation}
	\begin{aligned}
	T_{i}(X) = F_i \cdot (W \times T_{i-1}(X) + b) + (1 - F_i) \cdot T_{i-1}(X). \\
	\end{aligned}
	\end{equation}
	Where $T_0(X) = X$. $X$ denotes the input feature map of the directional convolution, $W$ and $b$ denote the shared weights and bias in iterative convolutions.

\begin{figure}
		\centering
		\includegraphics[page=1,width=0.42\textwidth]{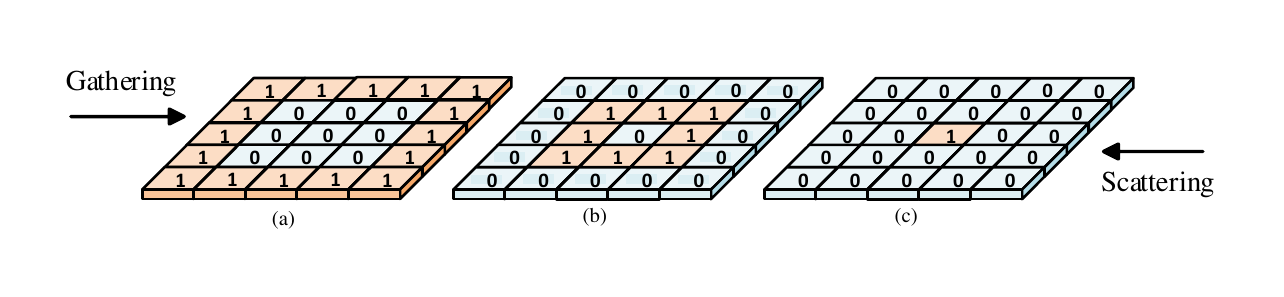}
		\caption{\label{characteristic}\small{\textbf{Characteristic Function. } A set of example characteristic functions used in Gathering and Scattering convolution with grid size 5 is illustrated. (a) denotes $F_1^{Gather}$ or $F_3^{Scatter}$, (b) denotes $F_2^{Gather}$ or $F_2^{Scatter}$, (c) denotes $F_3^{Gather}$ or $F_1^{Scatter}$}.}
		\vspace{-0.5cm}
	\end{figure}

	To explicitly handle the task of skeleton and contour feature map alignment, we specialize a pair of symmetric directional convolution operators, namely Scattering and Gathering convolutions (see Fig.~\ref{characteristic}). As illustrated in Fig.~\ref{block_conv}(a), the Gathering and Scattering convolutions updates feature outside-in and inside-out respectively. Gathering and Scattering convolution on size $n$ grid consists of $\lceil n/2 \rceil$ iterations.  
	
	We have addressed the sub-region feature fusion task, here we introduce how to partition a set of feature maps spatially into sub-regions. To fully exploit the spatial information over different locations, we use several different partitions for a set of feature maps to capture the diversity. Directional convolution is conducted in parallel on these partitions (see Fig.~\ref{block_conv}(c)). All output blobs are merged in an adaptive manner to produce the final convolutional features. Taking the Gathering convolution with grid size 4 as an example, using points $(0,0), (0,2), (2,0), (2,2)$ as grid partition points, we give out 4 kinds of partitions (which form the partition set P = \{$p_1, p_2, p_3, p_4$\}) on the feature map. One example of grid size 4, feature map size 8 is illustrated in Fig.~\ref{block_conv}(b). We denote Gathering convolution on each partition as $G_{p_i}$. So that we have the final result $G$ of advanced Gathering convolution to be: 
	\begin{equation}
	[W_{p_1},...,W_{p_n}] = \sigma (W \cdot [G_{p_1}(X),...,G_{p_n}(X)] + b).
	\end{equation}
	\begin{equation}
	G = \sum_{p_i \in P} W_{p_i} \cdot G_{p_i}(X).
	\end{equation}
	where $\sigma$ represents the sigmoid function, $[.]$ represents concatenation operation, $W_{p_i}$ is the estimated weight of feature from each partition. We reformulate the directional transform problem as the optimization procedure to path searching problem. Convolutions on different partitions represent different path that spatial transformation may take, and the weighted scheme represents a routing process among these paths. The illustration is provided in Fig.~\ref{block_conv}(c). The routing is learned from data, so that the Scattering and Gathering process between boundary and skeleton becomes possible. The output $G$ is latter used as input of the other branch.
	
	Directional convolution is a better alternative to normal convolution for its efficiency and flexibility. Normal convolution is good at feature extraction, but contains no specialized design for spatial transformation. However, in our directional convolution module, iterative convolutions are designed to be directional, which satisfy the needs of setting the direction of message flow explicitly. Redundant computation and parameters will be saved during each iteration. Besides, convolution weights are shared in a directional convolution block. So that compared to normal convolutions, using the same amount of parameters, the module can achieve much larger reception fields.
	
	\begin{figure*}
		\centering
		\includegraphics[page=1,width=0.82\textwidth]{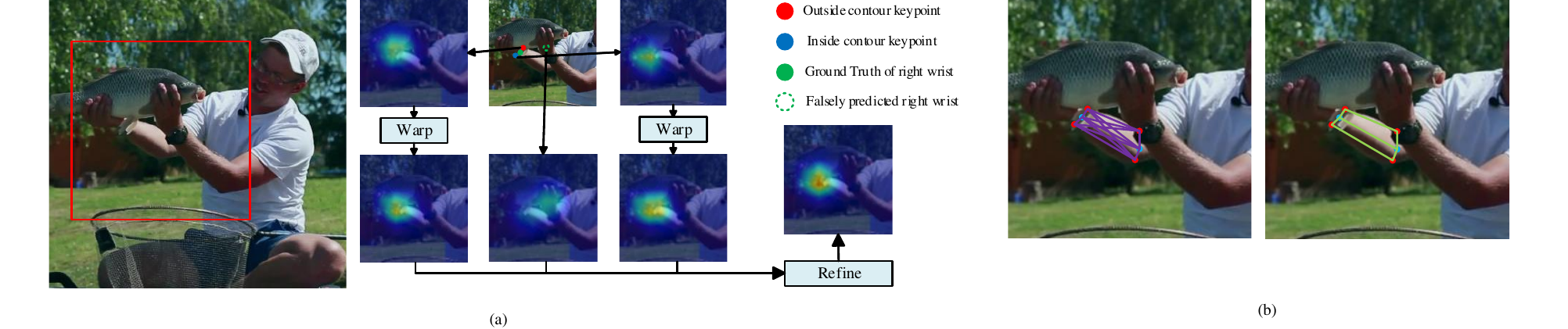}
		\caption{\label{pairwise_mapping}\small{\textbf{Pairwise mapping.} In (a), the coarse prediction of right wrist is wrong but later corrected by two rightly predicted landmarks on its sides. In (b), the difference between dense pairwise terms(left) and our important landmark pairs(right). Our definition remains rich structural information without losing simplicity.}}
		\vspace{-0.2cm}
	\end{figure*}

	\subsection{Pairwise Mapping}
	\label{sec_pairwise_mapping}
	
\begin{figure}[H] 
		\centering
		\includegraphics[page=1,width=0.42\textwidth]{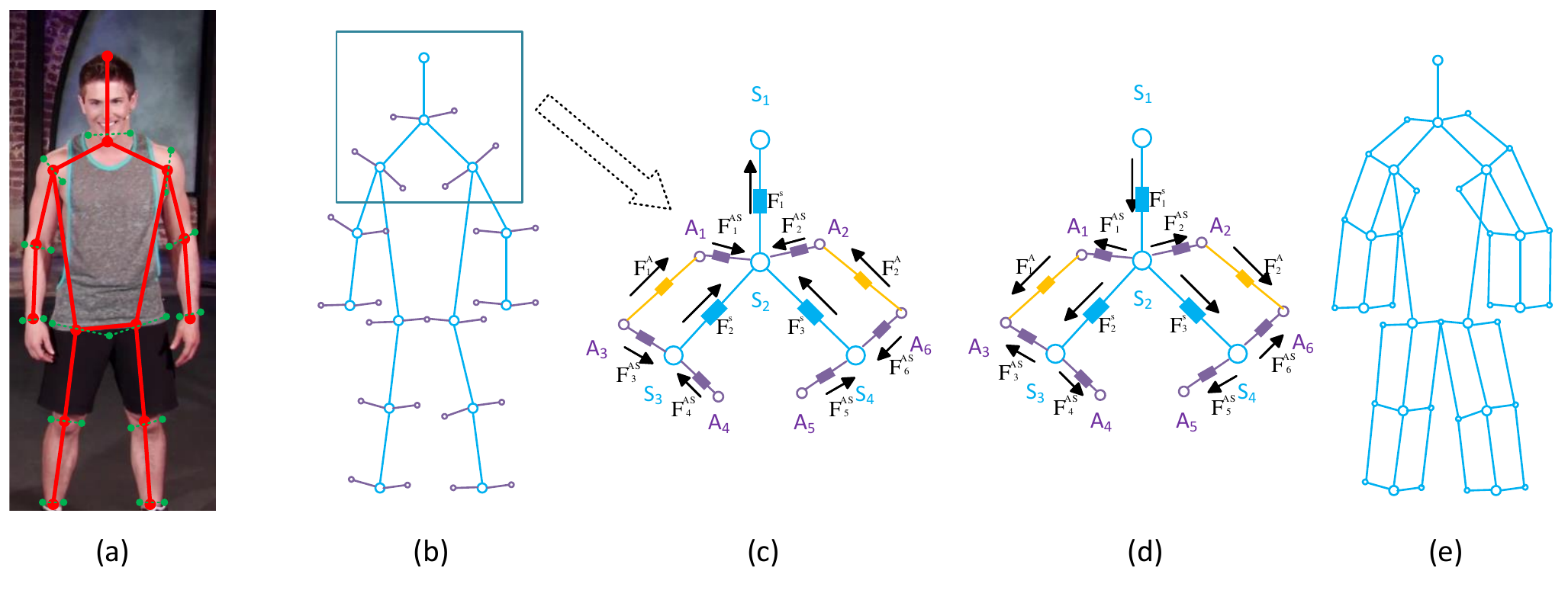}
		
		\caption{\label{i3}\small{\textbf{Message Passing.} (a) is a person image with annotated landmarks. (b) is the tree structured model of human with contour keypoints. (c,d) show message passing on part of the graph with different directions. (e) demonstrates message passing routes between pairwise body landmarks.}}
	\end{figure}

	To better preserve the consistency between skeleton and contour points, we propose the pairwise mapping between neighboring keypoint pairs. We first construct a graph to model the human body structure (see Fig.~\ref{i3}(e)). The nodes in the graph represent the skeleton and contour keypoints, and the edges represent the message passing routes. We design three types of joint relationship between: $(1)$ neighboring contour keypoints, $(2)$ skeleton keypoint and its neighbor contour keypoints, $(3)$ neighboring skeleton keypoints. We perform message passing along the edges in the graph as illustrated in Fig.~\ref{i3}(c)(d). Comparing to the work by Chu \textit{et al.}~\cite{chu2016structured}, we use feature warping to model the pairwise relationship between keypoints, instead of using directed kernels. Besides, novel pairwise relationship is included since we have taken contour points into consideration.
	
	\vspace{-0.3cm}
	\vspace{-0.3cm}
	
	To add pairwise mapping on the heatmaps of neighboring landmark pairs $(L_i, R_i)$, we tailor the feature warping strategy into our framework. First, in the middle of each stage, we add a new branch to estimate the bi-directional feature flows $Flow$ and $RevFlow$. The feature flow estimation network leverage features from contour and skeleton branches to estimate the mapping between them. Then, the estimated feature flow is used to warp the heatmaps from the source domain to the target domain (for example, $L_i$ is warped to the domain of $R_i$ by $Flow$). The warping operation is conducted in both directions. Ideally, after warping, the warped source heatmaps should be the same as the target heatmaps. To achieve this effect, a loss function is introduced to penalize the difference between the warped heatmaps and the target heatmaps, which we call pairwise loss. 
	This is formulated as:
	\begin{equation}
	\begin{aligned}
	\mathcal{L}_{P} &= \sum_{i=1}^n (||Warp(L_i, Flow_i) - R_i||^2  \\
	&+ ||Warp(R_i, RevFlow_i) - L_i||^2).
	\end{aligned}
	\end{equation}
	where $n$ denotes the number of pairs selected important keypoints. The predicted heatmaps for each pair is represented by $(L_i, R_i)$, 
	%and the warped points represented as $L_1, L_2,\cdots,L_n$ and the corresponding other keypoint represented as $R_1, R_2,\cdots, R_n$. 
	$Flow_i$ represents learned mapping from $L_i$ to $R_i$ while $RevFlow_i$ represents learned mapping from $R_i$ to $L_i$. The $Warp(H, Flow)$ function indicates the operation of warping heatmap $H$ with $Flow$. \footnote{The detailed definition of feature flow can be found in our supplementary material. }
	
	With the pairwise mapping loss, the network is encouraged to jointly learn keypoint locations and the mapping offsets between keypoint pairs and try to make consistent predictions. The warped heatmaps represent evidence of one keypoint location supported by other related keypoints. A fusion module is designed to combine the coarse heatmaps predicted by the network and the warped heatmaps produced by the warping module. In our implementation, the fusion model consists of two 1x1 convolution layers. By integrating evidence of locations, our network produce more accurate predictions. One example of the efficacy of pairwise mapping is demonstrated in Fig.~\ref{pairwise_mapping}.
	
	As shown in Fig.~\ref{framework}, our overall learning objective can be formulated as follows:
	\begin{equation}
	\label{formula6}
	\begin{aligned}
	\mathcal{L} = \sum_i \mathcal{L}_{stack_i} =\sum_i(\mathcal{L}_{S} + \mathcal{L}_{C} + \mathcal{L}_{P}),
	\end{aligned}
	\end{equation}
	where $L_S$, $L_C$, $L_P$ represent skeleton loss, contour loss and pairwise loss respectively.

	\section{Experiments}
	\subsection{Experiment Details}
	\label{details}
	For experiments on LSP and MPII, we adopt the two-stack net in~\cite{liu2018cascaded} as our baseline, except the complex CJN module in the original paper. For experiments on COCO, since~\cite{liu2018cascaded} didn't report results on COCO, we adopt a universal two-stack hourglass network~\cite{newell2016stacked} as our baseline. We perform the top-down multi-person pose estimation on COCO. A feature pyramid network~\cite{lin2017feature} based detector is used to generate human bounding box proposals (human detection AP is around 51), then the hourglass network is used to estimate human pose for each box proposal. 
	
	For data augmentation, images are cropped with the target person centered, and resized roughly to the same scale to fit for the 256x256 input size. While generating the training images, we randomly rotate (\textpm 40\textdegree) and flip the image. Random rescaling (0.7-1.3) is also performed. This data augmentation setting is consistent across all the experiments. More details are provided in the supplementary material. 
	
	\subsection{Results on TRB Estimation}
	We first evaluate the performance of several popular skeleton keypoint estimation approaches on the task of TRB Estimation, namely 4-stack hourglass~\cite{newell2016stacked}, Simple Baseline~\cite{xiao2018simple}, and Cascaded AIOI~\cite{liu2018cascaded}. Quantitative results on TRB estimation task are shown in Table~\ref{table_ctr_results}. We find that contour estimation is more challenging than skeleton estimation, resulting in lower keypoint accuracy. Our proposed TRB-Net outperforms all the state-of-the-art approaches, indicating its effectiveness of joint learning of skeleton keypoints and contour keypoints.
	
	\vspace{-2mm}
	
	\begin{table}[H] \label{tcr}
		\caption{Comparison with state of the art methods on MPII\_trb val.}
		\label{table_ctr_results}
		
		\centering 
		\resizebox{\columnwidth}{!}{%
			\begin{tabular}{llllllll|ll|ll}
				
				\toprule
				& {Head} & {Sho.} & {Elb.} & {Wri.} & {Hip} & {Knee}  & {Ank.} & Ske. & Con. & {Mean}\\
				
				\midrule
				Hourglass~\cite{newell2016stacked} & 96.8 & 95.2 & 89.2 & 85.2 & 87.4 & 83.9 & 81.5 & 89.0 & 85.3 & 86.6 \\
				Simple Baseline~\cite{xiao2018simple} Res-50 & 96.2 & 94.8 & 88.5 & 83.0 & 86.2 & 82.9 & 80.0 & 88.0 & 83.9 & 85.4\\ 
				Simple Baseline~\cite{xiao2018simple} Res-152 & 96.5 & 95.2 & 88.2 & 83.0 & 87.8 & 84.5 & 80.9 & 88.5 & 85.8 & 86.8\\
				Cascaded AIOI~\cite{liu2018cascaded} & 96.6 & 95.0 & 88.4 & 83.1 & 87.8 & 83.9 & 80.3 & 88.4 & 85.4 & 86.5 \\
				\midrule
				TRB-Net (Ours) & \textbf{97.1} & \textbf{95.6} & \textbf{90.2} & \textbf{85.6} & \textbf{89.3} & \textbf{86.4} & \textbf{83.5} & \textbf{90.1} & \textbf{87.2} & \textbf{88.2}\\
				\bottomrule
			\end{tabular}%
		}
	\end{table}
	
	\vspace{-4mm}
	
	\subsection{Ablation study}
	\label{ablation_part}
	To thoroughly investigate the efficacy of the proposed TRB and the message passing components, we conduct extensive ablation study on the MPII\_trb validation set.
	
	\textbf{Directional Convolution.} Comparing to the baseline which only uses skeleton keypoints (\textbf{Skeleton}) or contour keypoints (\textbf{Contour}) for training, the two branch network which jointly learns the skeleton and contour landmarks (\textbf{Multitask}) achieves better results. Based on that two branch model, we explored the effect of different techniques used to promote feature-level message passing between the skeleton branch and the contour branch. We show that adding the X-structured message passing unit (\textbf{Xs}) improves the prediction accuracy on both skeleton and contour keypoints. Then we found that Directional Convolution (\textbf{DC}) can be a better replacement due to its efficiency and flexibility, in our experiments, DC beats Xs by 0.6\% in mean PCKh of TRB. For further analysis on the efficacy of DC, we compare the DC unit and normal convolutions with the same parameter size. The results turns out that DC beats \textbf{Normal Conv} with a large margin of 0.9\%. We also remove the adaptive weight in multi-path directional convolution fusion (\textbf{DC-Ada}) and it leads to 0.3\% drop. It shows that the adaptive weighting scheme is important for DC to work. DC improved 1.4\% and 2.1\% over the baseline for skeleton and contour keypoints respectively, which indicates the effectiveness of our proposed message passing scheme. Table.~\ref{table3} presents ablation results of different message passing schemes discussed above.
	
	\begin{table}[H] \label{t3}
		\caption{Ablation study on directed convolution}
		\label{table3}
		
		\centering 
		\resizebox{\columnwidth}{!}{%
			\begin{tabular}{l|llllllll}
				
				\toprule
				Acc\textbackslash{}Approach & Skeleton & Contour & Multitask & Xs & Normal Conv & DC - Ada & DC\\
				
				\midrule
				Ske. & 88.0 & - & 88.6 & 88.9 & 88.7 & 89.1 & \textbf{89.4}\\
				Con. &  - & 84.1 & 84.8 & 85.6 & 85.3 & 85.9 & \textbf{86.2}\\
				Mean & - & - & 86.2 & 86.8 & 86.5 & 87.1 & \textbf{87.4} \\
				\bottomrule
			\end{tabular}%
		}
	\end{table}

	\textbf{Pairwise Mapping.} The pairwise mapping strategy further enforces explicit triplet representation consistency. By adding pairwise mapping module to the Multitask baseline with contour annotations (\textbf{Contour}), we obtain 1.0\% improvement in mean PCKh. We further demonstrate the efficacy of integrating pairwise mapping into pose estimation by examining intermediate results. At each stack, we first get the coarse estimation (\textbf{-c}). Then, we warp them with data learned warping to enforce representation consistency. Finally, we fuse the original heatmaps and warped heatmaps to generate our finer estimation results (\textbf{-f}). We find that pairwise mapping and fusion consistently improve the TRB estimation results. The TRB estimation results after pairwise mapping and fusion are consistently better than the original coarse estimation, the improvement being extremely large in the early stack of the network. By combining directional convolution with pairwise mapping (\textbf{DC + PM}), the overall performance is further boosted to 87.6\%. Detailed results are listed in Table~\ref{table4}.
	
	\begin{figure}
		\centering
		\includegraphics[page=1,width=0.42\textwidth]{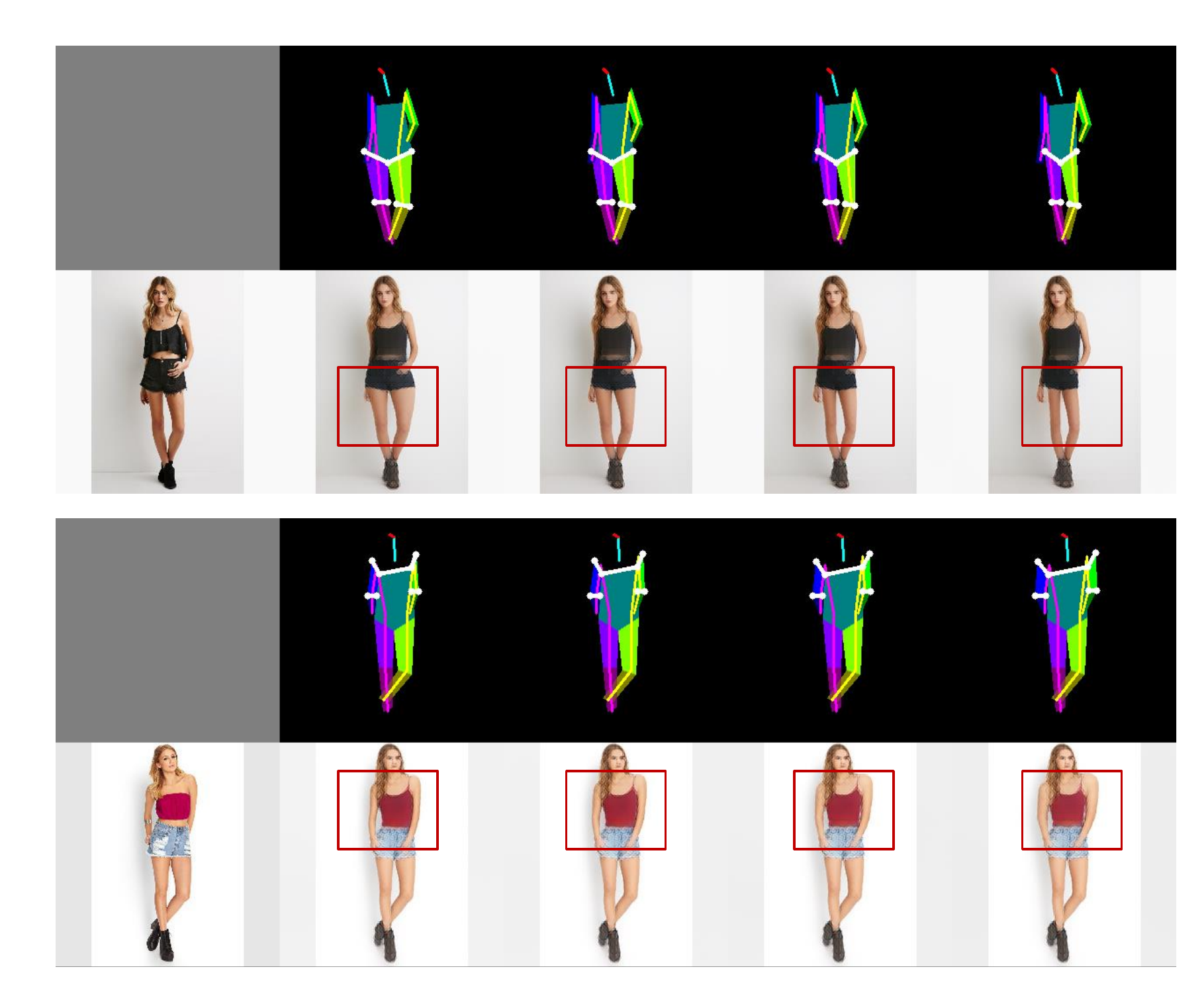}   
		\caption{\label{slimmer}\small\textbf{Image Generation based on contour points}. We edit contour points of upper legs and upper arms to generate human with different body shape. The 1st and the 3rd rows denotes the appearance we used for generation. First line in strong to slim order, third line in slim to strong order. The edited contour keypoints are highlighted in white. The 2nd and the 4th rows denote the generated images.}
	\end{figure}

	\begin{table}[H] \label{t4}
		\caption{Ablation study on Pairwise Mapping}
		\label{table4}
		
		\centering 
		\resizebox{\columnwidth}{!}{%
			\begin{tabular}{l|llll|llll}
				\toprule
				Acc\textbackslash{}Approach & stack1-c & stack1-f & stack2-c & stack2-f & Multitask & PM & DC + PM\\
				\midrule
				Ske. & 86.2 & 87.2 & 88.9 & 89.1 & 88.6 & 89.2 & \textbf{89.6}\\
				Con. & 83.2 & 84.2 & 85.4 & 86.1 & 84.8 & 86.1 & \textbf{86.5}\\
				Mean & 84.3 & 85.3 & 86.6 & 87.2 & 86.2 & 87.2 & \textbf{87.6} \\
				\bottomrule
			\end{tabular}%
		}
	\end{table}

	\subsection{TRB for Shape Editing}
	TRB contains rich human shape information, which can be exploited in various applications. The prospect of using TRB for conditional image generation and photo editing is promising. In this section, we demonstrate its application on human shape editing. Following~\cite{esser2018variational}, we develop a variational u-net for human generation and style-transfer conditioned on human pose and shape. We conduct experiments on DeepFashion~\cite{liuLQWTcvpr16DeepFashion}. Some results are displayed in Fig.~\ref{slimmer}, in which we edit contour points to change the shape of the upper leg and upper body while keeping the pose and appearance fixed. Specifically, when editting leg shape, we move the medial contour point along the axis defined by two contour point in the same triplet. Besides that, when generating stronger upper body, the distance between two lateral shoulder contour points is increased. 
	
	TRB is a compact and powerful shape representation, which makes human shape editing possible given only a handful of semantic keypoints. Semantic parsing provides pixel-level human part information. However, due to the lack of accurate localization information, it cannot be directly used for shape editing. 3D representation like DensePose~\cite{alp2018densepose} can be used for shape editing~\cite{neverova2018dense}, but they do not support arbitrary 2D shape manipulation. Comparing to DensePose, our editing is much lighter, with no need of heavy intermediate representation.

	\subsection{Results on Skeleton Estimation Datasets}
	
	\begin{table}[h!]\label{t1}
		\caption{Quantitative results on LSP test set (PCK@0.2)}
		\label{sample-table}
		
		\centering 
		\resizebox{\columnwidth}{!}{%
			\begin{tabular}{llllllll|ll}
				\toprule
				&Head & Shoulder & Elbow & Wrist & Hip & Knee  & Ankle & Mean & AUC\\ \midrule
				Chu et al. CVPR'17~\cite{chu2017multi}& 98.1  & 93.7  & 89.3  & 86.9  & 93.4  & 94.0 & 92.5 & 92.6 & 64.9 \\
				Yang et al. ICCV'17~\cite{yang2017learning}& 98.3  & 94.5  & 92.2  & 88.9  & 94.4  & 95.0 & 93.7 & 93.9 & 68.5\\
				Ning et al. TMM'17~\cite{ning2018knowledge} & 98.2  & 94.4  & 91.8  & 89.3  & 94.7  & 95.0 & 93.5 & 93.9 & 69.1\\
				Chou et al. arxiv'17~\cite{chou2017self} & 	98.2& 94.9 & 92.2 &	89.5 & 94.2 & 95.0 & 94.1 &	94.0 & 69.6 \\
				Zhang et al. arxiv'19~\cite{zhang2019human} & 98.4 & 94.8 & 92.0 & 89.4 & 94.4 & 94.8 & 93.8 & 94.0 & - \\
				\midrule
				Liu et al. AAAI'18~\cite{liu2018cascaded} & 98.1 & 94.0 & 91.0 & 89.0 & 93.4 & 95.2 & 94.4  & 93.6 & - \\
				Ours &  98.5 & 95.3 & 92.6 & 90.6 & 93.8 & 95.8 & 95.5 & 94.5 & 69.9\\ 
				\bottomrule
			\end{tabular}%
		}
		
	\end{table}
	
	\textbf{LSP.} 
	Table~\ref{sample-table} presents experimental results of our approach and previous methods on LSP dataset. Following the setting of~\cite{yang2013articulated} , the Percentage Correct Keypoints (PCK) metric is used for evaluation, in which the standard distance is 0.2 times the distance between the person's left shoulder and the right hip on the image. Our approach achieves 94.5\% PCK and consistently outperforms the state-of-the-arts. In particular, our method surpasses previous methods on hard keypoints like wrist and ankles by considerable margins. Our success on part localization confirms the benefit of the boundary information around limbs.
	
	\begin{table}[H] \label{t2}
		
		\caption{Quantitative results on MPII test set (PCKh@0.5)}
		\label{sample-table1}
		\centering 
		\resizebox{\columnwidth}{!}{%
			\begin{tabular}{llllllll|ll}
				\toprule
				&Head & Shoulder & Elbow & Wrist & Hip & Knee  & Ankle & Mean & AUC \\
				\midrule
				Ning et al., TMM'17~\cite{ning2018knowledge} & 98.1  & 96.3  & 92.2  & 87.8  & 90.6  & 87.6 & 82.7 & 91.2 & 63.6 \\
				Chu et al., CVPR'17~\cite{chu2017multi} & 98.5 & 96.3  & 91.9  & 88.1  & 90.6  & 88.0 & 85.0 & 91.5 & 63.8 \\
				Nie et al., CVPR'18~\cite{Nie_2018_CVPR} & 98.6  & 96.9  & 93.0  & 89.1  & 91.7  & 89.0 & 86.2 & 92.4 & 65.9 \\
				Zhang et al. arxiv'19~\cite{zhang2019human} & 98.6 & 97.0 & 92.8 & 88.8 & 91.7 & 89.8 & 86.6 & 92.5 & - \\
				
				\midrule
				Liu et al. AAAI'18~\cite{liu2018cascaded} & 98.4 & 96.4 & 92.0 & 87.8 & 90.7 & 88.3 & 85.3 & 91.6 & 64.6 \\
				Ours&  98.5  & 96.6  & 92.6 & 88.3  & 91.6 & 89.2 & 86.5 & 92.2 & 65.4 \\
				\bottomrule
			\end{tabular}%
		}
		
	\end{table}
	
	\textbf{MPII.}
	Table~\ref{sample-table1} presents results on MPII dataset. PCKh was chosen as the measurement, following~\cite{andriluka20142d}. Under this metric, the threshold of distance to measure the accuracy is half of the head size. Note that Nie et al. used additional dataset LIP~\cite{gong2017look} for training, which contains 50000 images with pixel-wise annotated semantic human part labels, and Zhang et al. used additional dataset LSP (which contains more challenging poses) for training. By exploiting the visual evidence on human contour, our model outperforms the state-of-the-art methods which only use MPII for training, and is competitive to methods using external data.

\begin{figure}
		\centering
		\includegraphics[page=1,width=0.42\textwidth]{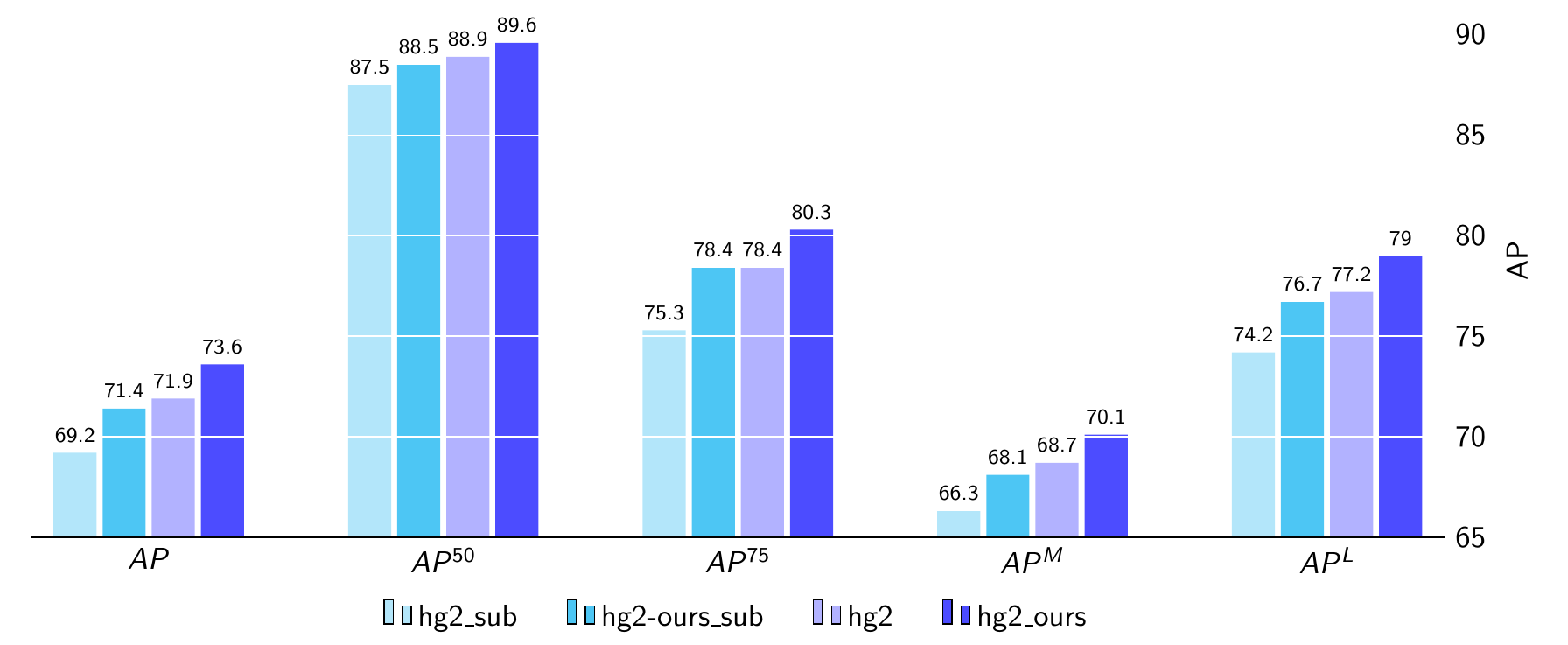}   
		\caption{\small \label{coco_ap}\textbf{Results on COCO validation.} 'sub' denotes using only half of the data for training. The results are obtained with single-scale testing and flipping.}
	\end{figure}

	\textbf{COCO.}
	Half of 150000 human instances in COCO was annotated with TRB. We follow settings in~\cite{li2019rethinking} to conduct experiments. Our baseline is finely tuned, with higher accuracy comparing to results reported in~\cite{li2019rethinking} (71.9 v.s. 70.9). Fig.\ref{coco_ap} shows that considerable and consistent improvement is made beyond the strong baseline. With half of the data, our method (hg2-ours\_sub) reached competitive performance comparing to the baseline using all data, which illustrated the efficiency of the contour keypoints we designed. Please refer to supplementary materials for results on COCO test-dev.
	
	\section{Conclusion}
	In this work, we propose TRB, a new body representation including both 2D human pose and shape. Contour keypoints are included as a compact representation of 2D shape beyond traditional skeleton landmarks. We set a benchmark for the newly proposed TRB estimation task, comparing different 2D pose estimation approaches in the new setting. We further propose an effective multi-task network to learn human skeleton and contour jointly. Using TRB based conditional human image generation, we illustrate the effectiveness and explicitness of the proposed representation. 
	%------------------------------------------------------------------------
	
	{\small
		\bibliographystyle{ieee_fullname}
		\bibliography{egbib}
	}
	
\end{document}